\documentclass{article} 
\usepackage{iclr2015,times}
\usepackage{hyperref}
\usepackage{url}
\usepackage[dvips]{graphicx}
\usepackage[latin1]{inputenc}
\usepackage[export]{adjustbox}
\usepackage{amssymb,amsmath,array}
\usepackage{caption}
\iclrconference
\title{Learning Temporal Dependencies in Data Using a DBN-BLSTM}

\author{
Kratarth Goel \\
Department of Computer Science\\
BITS Pilani\\
Goa, India \\
\texttt{kratarthgoel@gmail.com} \\
\And
Raunaq Vohra \\
Department of Mathematics\\
BITS Pilani\\
Goa, India \\
\texttt{ronvohra@gmail.com}
}

\begin{document}

\maketitle

\begin{abstract}

Since the advent of deep learning, it has been used to solve various problems using many different architectures. The application of such deep architectures to auditory data is also not uncommon. However, these architectures do not always adequately consider the temporal dependencies in data. We thus propose a new generic architecture called the  Deep Belief Network - Bidirectional Long Short-Term Memory (DBN-BLSTM) network that models sequences by keeping track of the temporal information while enabling deep representations in the data. We demonstrate this new architecture by applying it to the task of music generation and obtain state-of-the-art results.

\end{abstract}

\section{Introduction}
\label{intro}

Deep architectures have been integral in creating a paradigm shift in the way we tackle most pattern recognition problems today. Deep belief networks (DBNs), for instance, are designed to maximize the variational lower bound of the log-likelihood, by hierarchical representation of data. They do not take into account the temporal information contained in speech signals, for example. The same goes for various other approcahes that have tried to use deep architectures for speech recognition (as explained in \cite{A04,A05}).  

Given that music is an inherently dynamic, it seems natural to consider recurrent neural networks (RNNs) as a good baseline technique for music generation. However, these models do not perform as well as deep networks (\cite{A08}). Another alternative can be to train RNNs 'end-to-end' rather than combining them with Hidden Markov Models (HMMs). The results achieved in \cite{A06} clearly support our claim about the importance of modeling temporal dependencies in sequential data.

RNNs are inherently deep in time, and this property is used when we train them using Backpropogation Through Time (BPTT). We believe that an amalgamation of the ability of RNNs to learn temporal dependencies in the data with the depth provided by a DBN would lead to more expressive RNNs - or their more sophisticated counterparts, Long Short-Term Memory (LSTM) networks - which are capable of modeling long term temporal dependencies. Essentially, we need an architecture that models the temporal nature of auditory data but at the same time also takes advantage of the hierarchical representations that result from the use of deep belief networks. As such, we introduce a deep architecture - the DBN-BLSTM - which gives state of the art results on the generative task of polyphonic music generation.

\section{DBN}

\emph{Restricted Boltzmann Machines} (RBMs) are energy based models with their energy function $E(v,h)$ defined as:

\begin{equation}
E(v,h) = - b_v'v - b_h'h - h'Wv
\end{equation}

where $W$ represents the weights connecting the units of the visible $(v)$ and hidden $(h)$ layers and $b_v$, $b_h$ are the biases of the visible and hidden layers respectively.

Samples can be obtained from a RBM by performing block Gibbs sampling, where visible units are sampled simultaneously given fixed values of the hidden units. Similarly, hidden units are sampled simultaneously given the visible unit values. A single step in the Markov chain is thus taken as follows: 

\begin{equation}
\begin{split}
h^{(n+1)} = \sigma(W'v^{(n)} + b_h) \\
v^{(n+1)} = \sigma(W h^{(n+1)} + b_v),
\end{split}
\end{equation}

where $\sigma$ represents the sigmoid function acting on the activations of the $(n + 1)^{th}$ hidden and visible units. Several algorithms have been devised for RBMs in order to efficiently sample from $p(v,h)$ during the learning process, the most effective being the well-known \emph{contrastive divergence} (CD-$k$) algorithm (\cite{A10}).

RBMs can be stacked and trained greedily to form \emph{Deep Belief Networks} (DBNs). DBNs are graphical models which learn to extract a deep hierarchical representation of the training data (\cite{A11}). They model the joint distribution between the observed vector $\mathbf{v}$ and the $\ell$ hidden layers $h^k$ as follows:

\begin{equation}
P(\mathbf{v}, h^1, \ldots, h^{\ell}) = \left(\prod_{k=0}^{\ell-2} P(h^k|h^{k+1})\right) P(h^{\ell-1},h^{\ell})
\end{equation}

where $\mathbf{v}=h^0$, and  $P(h^{k-1} | h^k)$ is a conditional distribution for the visible units conditioned on the hidden units of the RBM at level $k$, and $P(h^{\ell-1}, h^{\ell})$ is the visible-hidden joint distribution in the top-level RBM. 

\section{RNN and BLSTM} 

\begin{figure}[h!]
\centering
\includegraphics[scale=0.5]{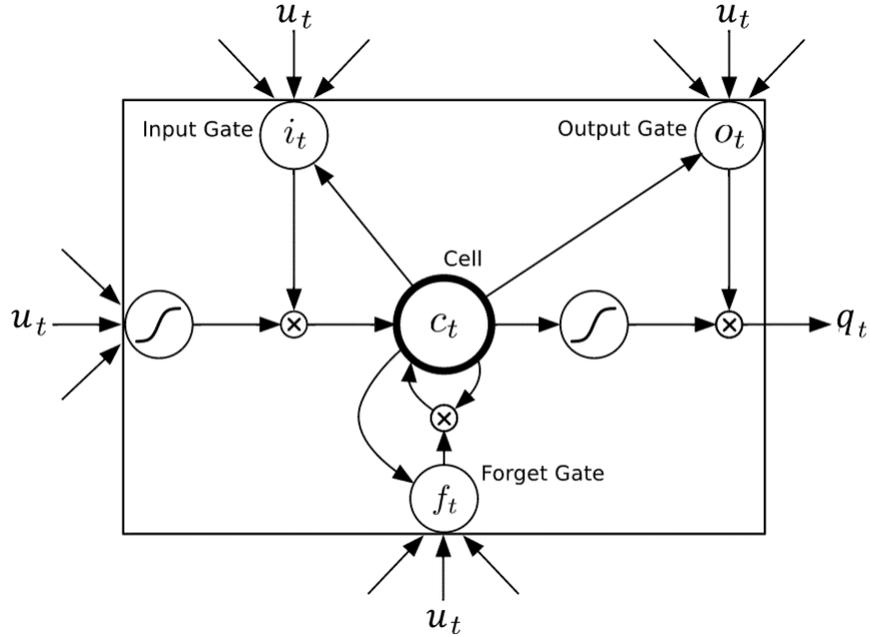}
\caption{A LSTM cell.}\label{LSTM}
\end{figure}

A \emph{Recurrent Neural Network} (RNN) is different from a standard network in that it takes a sequence $\mathbf{v} = (v_1, v_2, . . . , v_T)$ as input, and iterates over it from $t = 1$ to $T$, to produce the following:

\begin{equation}
{q_t} = \Phi (b_{{q_t}} + \overrightarrow{W_{vq}}v_t + \overrightarrow{W_{qq}}{q_{t-1}})
\end{equation}

where $\mathbf{q} = (q_1, q_2, . . . , q_T)$ is a vector representing the hidden unit. The $b$ terms are bias vectors (eg. $b_q$ represents bias of the hidden layer). The function $\Phi$ is usually the application of elementwise sigmoid ($\sigma$), in the case of a general RNN as can be seen from Eq. (5). However, when a LSTM is being used, we define $\Phi$ as follows:

\begin{equation}
u_t = \sigma (b_u + W_{vu}v_t + W_{uu}u_{t-1})
\end{equation}
\begin{equation}
i_t = \sigma(b_i + W_{ci}c_{t-1} + W_{qi}q_{t-1} + W_{ui}u_t)
\end{equation} 
\begin{equation}
f_t = \sigma(b_f + W_{cf}c_{t-1} + W_{qf}q_{t-1} + W_{uf}u_t)
\end{equation} 
\begin{equation}
c_t = f_tc_{t-1} + i_t\sigma(W_{uc}u_t + W_{qc}q_{t-1} + b_c)
\end{equation} 
\begin{equation}
o_t = \sigma(b_o + W_{co}c_{t} + W_{qo}q{t-1} + W_{uo}u_t)
\end{equation} 
\begin{equation}
q_t = o_ttanh(c_t)
\end{equation} 

where the $W$ terms denote the weight martrices (eg. $W_{vq}$ is the weight matrix between the visible and the hidden unit). Further, $u$ represents the input of the LSTM memory cell and $i$, $f$, $o$ and $c$ are the \emph{input} gate, \emph{forget} gate, \emph{output} gate and \emph{cell} activation vectors respectively, all of which are of the same size as the hidden vector $q$.  LSTM memory cells are good at finding and exploiting long range context. Figure \ref{LSTM} illustrates a single LSTM memory cell.

The basic idea of the Bidirectional LSTM (BLSTM) is to present each training sequence forwards and backwards to two separate LSTMs, both of which are connected to the same output layer. This means that for every point in a given sequence, the BLSTM has complete, sequential information about all points before and after it. Also, because the net is free to use as much or as little of this context as necessary, there is no need to find a (task-dependent) time-window or target delay size.

The forward hidden sequence is calculated as specified in Eq. (5). The backward hidden sequence $\overleftarrow{q_t}$  of the bidirectional LSTM $r$ is calculated by iterating over $v$ from t = T to 1, as follows:

\begin{equation}
\overleftarrow{q_t} = \Phi (b_{\overleftarrow{q_t}} + \overleftarrow{W_{vq}}v_t + \overleftarrow{W_{qq}}\overleftarrow{q_{t-1}})
\end{equation} 

\section{The DBN-BLSTM}

\subsection{The Architecture}

\begin{figure}[h!]
\centering
\includegraphics[scale=0.5]{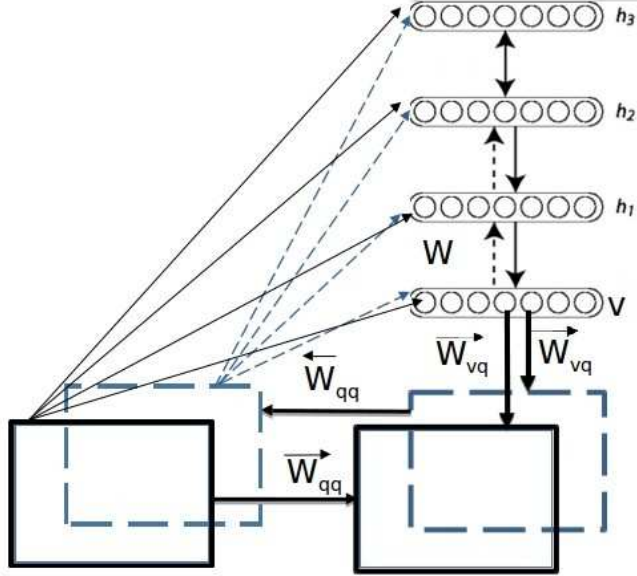}
\caption{The DBN-BLSTM at time instants $t_0$ and $t_1$. The solid rectangular boxes represent the forward chain LSTM and the dashed rectangular boxes represent the backward chain LSTM. The arrows from the LSTM units to the layers of the DBN represent the biases of the respective units (For example, the arrow connecting the solid rectangular box to the visible layer of the DBN represents $b_{v_0}$). The same structure can be extended for the length of the enitre sequence being modeled.}\label{DBN-BLSTM}
\end{figure}

The DBN-BLSTM is an extension of the generative model (RNN-DBN) proposed by \cite{A09}. There are a few significant improvements made to this model, most notably the replacement of the RNN with a LSTM, which is a more powerful neural architecture capable of modeling temporal dependencies across large time steps. This ensures the model retains information about the sequence generated for a longer time duration. Since we are discussing generative models, this property lends itself exceptionally well to modeling creativity, for instance to music generation, where choosing an LSTM over a RNN would result in less repetition and more varied music due to better generalization, since the improved memory of the LSTM would possess more information about previously generated music in the sequence as compared to a RNN. The network is illustrated in Figure \ref{DBN-BLSTM}.

There are essentially two points for interaction between the DBN and the BLSTM. The input to the BLSTMs are calculated as shown in Eq. (4) and (11) (for the forward and the backward LSTMs respectively) where the $v$ is the visible layer of the DBN. The second point of interaction between the two architectures is where the biases of the visible and hidden units of the DBN are calculated as follows:

\begin{equation} 
        b_{v_t} = b_{v_0} + W_{uv}u_{t-1} + W_{qv}q_{t-1}
\end{equation}
\begin{equation}
        b_{h^{(n)}_t} = b_h + W_{uh^{(n)}}u_{t-1} + W_{qh^{(n)}}q_{t-1}
\end{equation}

where $b_{h^{(n)}_t}$ represents the bias vector for the $n$\emph{\textsuperscript{th}} hidden layer, and $b_{v_t}$ the bias of the visible layer at the $t\textsuperscript{th}$ time step of the recurrence for the DBN. Thus the temporal dependencies learned by the BLSTM strongly influences the hierarchical representations that are learned by the DBN to construct the joint probability distribution $P(\mathbf{v}, h^1, \ldots, h^{\ell})$ given by Eq. (3).

\subsection{Training the Network}

At each time step $t$, the input $v_t$ is provided to the DBN. The corresponding input $u_t$ to the BLSTM is provided using Eq. (5). The biases to the layers of the DBN are calculated as given by Eq. (12) and (13), using the $\Phi$ function of the BSLTM given by Eq. (4) and (11). These biases are then used for calculating the joint probability distribution between the observed and hidden layers of the DBN given in Eq. (3). For sampling from the DBN we use Eq. (2). 

It must be noted here that the input $v_t$ to the DBN is a binary vector at each time step $t$, and the model for the music generation task is thus composed of only binary layers. However, this is not a limitation of our technique, and the model can easily be extended to handle real valued data as well. One such possible extension could have the visible layer of the DBN composed of Gaussian units and the subsequent layers could be ReLU units with dropout incorporated per layer.

\section{Experiment and Results}

We demonstrate our technique by applying it to the task of polyphonic music generation. We used a DBN-BLSTM with 3 hidden DBN layers - each having 150 binary units - and 150 binary units in the BLSTM. The visible layer has 88 binary units, corresponding to the full range of the piano from A0 to C8. Dropout is incorporated in each layer. We used our technique on four datasets - \emph{JSB Chorales} , \emph{MuseData}\footnotemark[1]\footnotetext[1]{http://www.musedata.org}, \emph{Nottingham}\footnotemark[2]\footnotetext[2]{ifdo.ca/~seymour/nottingham/nottingham.html} and \emph{Piano-midi.de}.  Only raw MIDI data has been given as input to the DBN-BLSTM. We evaluate our models qualitatively by generating sample sequences and quantitatively by using the \emph{log-likelihood} (LL) as a performance measure. Results (some of which are reproduced from \cite{A12}) are presented in Table 1.

\captionof{table}{Log-likelihood (LL) for various musical models in the polyphonic music generation task.}

	\begin{tabular}{|p{3cm}|p{1.45cm}|p{1.65cm}|p{2.05cm}|p{1.45cm}|}
	

		\hline
		\bfseries{Model} & \bfseries{JSB Chorales (LL)} & \bfseries{MuseData (LL)} & \bfseries{Nottingham (LL)} & \bfseries{Piano-Midi.de (LL)}\\
		\hline
		Random & -61.00 & -61.00 & -61.00 & -61.00\\
		RBM & -7.43 & -9.56 & -5.25 & -10.17\\
		NADE & -7.19 & -10.06 & -5.48 & -10.28\\
		RNN-RBM & -7.27 & -9.31 & -4.72 & -9.89\\	
		RNN (HF) & -8.58 & -7.19 & -3.89 & -7.66\\	
		RNN-RBM (HF) & -6.27 & -6.01 & -2.39 & -7.09\\	
		RNN-DBN & -5.68 & -6.28 & -2.54 & -7.15 \\
		RNN-NADE (HF) & -5.56 & -5.60 & -2.31 & -7.05\\

				\hline
		\bfseries{DBN-BLSTM} & \bfseries{-3.47} & \bfseries{-3.91} & \bfseries{-1.32} & \bfseries{-4.63}\\

		\hline
	
	\end{tabular}

\vspace*{0.5cm}

The results clearly indicate that our technique performs significantly better than the current state-of-the-art. 

\section{Conclusions and Future Work}

We have proposed a generic technique called DBN-BLSTM for modeling sequences and have demonstrated its successful application to polyphonic music generation. We have used four datasets for evaluating our technique and have obtained state-of-the-art results. In the future, we look to work on other powerful architectures that help learning temporal dependencies in data without compromising on the powerful hierarchical representations that DBNs provide.

\section{Acknowledgements}

We would like to thank Yoshua Bengio for helpful discussions. We would also like to thank the developers of Theano (\cite{A01}; \cite{A02}), which we have used for all our experiments.

\bibliography{example_paper}

\begin{thebibliography}{10}
\providecommand{\natexlab}[1]{#1}
\providecommand{\url}[1]{\texttt{#1}}
\expandafter\ifx\csname urlstyle\endcsname\relax
  \providecommand{\doi}[1]{doi: #1}\else
  \providecommand{\doi}{doi: \begingroup \urlstyle{rm}\Url}\fi

\bibitem[Bastien et~al.(2012)Bastien, Lamblin, Pascanu, Bergstra, Goodfellow,
  Bergeron, Bouchard, and Bengio]{A01}
Bastien, Fr{\'{e}}d{\'{e}}ric, Lamblin, Pascal, Pascanu, Razvan, Bergstra,
  James, Goodfellow, Ian~J., Bergeron, Arnaud, Bouchard, Nicolas, and Bengio,
  Yoshua.
\newblock Theano: new features and speed improvements, 2012.

\bibitem[Bergstra et~al.(2010)Bergstra, Breuleux, Bastien, Lamblin, Pascanu,
  Desjardins, Turian, Warde-Farley, and Bengio]{A02}
Bergstra, James, Breuleux, Olivier, Bastien, Fr{\'{e}}d{\'{e}}ric, Lamblin,
  Pascal, Pascanu, Razvan, Desjardins, Guillaume, Turian, Joseph, Warde-Farley,
  David, and Bengio, Yoshua.
\newblock Theano: a {CPU} and {GPU} math expression compiler.
\newblock In \emph{Proceedings of the Python for Scientific Computing
  Conference ({SciPy})}, June 2010.

\bibitem[Boulanger-Lewandowski et~al.(2012)Boulanger-Lewandowski, Bengio, and
  Vincent]{A12}
Boulanger-Lewandowski, Nicolas, Bengio, Yoshua, and Vincent, Pascal.
\newblock Modeling temporal dependencies in high-dimensional sequences:
  Application to polyphonic music generation and transcription.
\newblock In \emph{ICML}, 2012.

\bibitem[Deng \& Yu(2011)Deng and Yu]{A05}
Deng, Li and Yu, Dong.
\newblock Deep convex network: A scalable architecture for speech pattern
  classification.
\newblock In \emph{Interspeech}. International Speech Communication
  Association, August 2011.

\bibitem[Goel et~al.(2014)Goel, Vohra, and Sahoo]{A09}
Goel, Kratarth, Vohra, Raunaq, and Sahoo, J.~K.
\newblock Polyphonic music generation by modeling temporal dependencies using a
  {RNN-DBN}.
\newblock In \emph{ICANN}, pp.\  217--224, 2014.

\bibitem[Graves et~al.(2013)Graves, Mohamed, and Hinton]{A06}
Graves, Alex, Mohamed, Abdel{-}rahman, and Hinton, Geoffrey~E.
\newblock Speech recognition with deep recurrent neural networks.
\newblock \emph{CoRR}, abs/1303.5778, 2013.

\bibitem[Hinton(2002)]{A10}
Hinton, Geoffrey~E.
\newblock Training products of experts by minimizing contrastive divergence.
\newblock \emph{Neural Computation}, 14\penalty0 (8):\penalty0 1771--1800,
  2002.

\bibitem[Hinton \& Osindero(2006)Hinton and Osindero]{A11}
Hinton, Geoffrey~E. and Osindero, Simon.
\newblock A fast learning algorithm for deep belief nets.
\newblock \emph{Neural Computation}, 18, 2006.

\bibitem[Hinton et~al.(2012)Hinton, Deng, Yu, Dahl, Mohamed, Jaitly, Senior,
  Vanhoucke, Nguyen, Sainath, and Kingsbury]{A04}
Hinton, Geoffrey~E., Deng, Li, Yu, Dong, Dahl, George~E., Mohamed,
  Abdel{-}rahman, Jaitly, Navdeep, Senior, Andrew, Vanhoucke, Vincent, Nguyen,
  Patrick, Sainath, Tara~N., and Kingsbury, Brian.
\newblock Deep neural networks for acoustic modeling in speech recognition: The
  shared views of four research groups.
\newblock \emph{Signal Processing Magazine, IEEE}, 29\penalty0 (6):\penalty0
  82--97, Nov 2012.

\bibitem[Vinyals et~al.(2012)Vinyals, Ravuri, and Povey]{A08}
Vinyals, Oriol, Ravuri, Suman, and Povey, Daniel.
\newblock {R}evisiting recurrent neural networks for robust {ASR}.
\newblock IEEE International Confrence on Acoustics, Speech, and Signal
  Processing (ICASSP), March 2012.

\end{thebibliography}
\bibliographystyle{iclr2015}

\end{document}